\documentclass[11pt]{article}

\usepackage[preprint]{acl}

\usepackage{times}
\usepackage{latexsym}

\usepackage[T1]{fontenc}

\usepackage[utf8]{inputenc}

\usepackage{microtype}

\usepackage{inconsolata}

\usepackage{graphicx}
\usepackage{booktabs}
\usepackage{amsmath}
\title{Light or Full Verb? A Minimal-Pair Dataset for Probing Phraseological Competence in Language Models}

\author{
 \textbf{Francesca Franzon\textsuperscript{1}},
 \textbf{Nicolas Rosàs Gómez\textsuperscript{1}},
 \textbf{Leo Wanner\textsuperscript{1}$^,$\textsuperscript{2}}
\\
 \textsuperscript{1}Universitat Pompeu Fabra (UPF)\\
 \textsuperscript{2}Catalan Institute for Research and Advanced Studies (ICREA)
\\
  \href{mailto:francesca.franzon@upf.edu}{francesca.franzon@upf.edu}, \href{mailto:nicolas.rosas@upf.edu}{nicolas.rosas@upf.edu}, \href{mailto:leo.wanner@upf.edu}{leo.wanner@upf.edu}
 }

\begin{document}
\maketitle

\begin{abstract}
Frequent verbs such as \emph{have} and \emph{make} can function either as collocates in light-verb constructions or as full lexical predicates, as in \emph{make a decision} vs. \emph{make a cake}. Whether language models represent this distinction, and whether such representations vary across languages, remains unclear. We introduce a large-scale controlled dataset in English, Spanish, and French, comprising minimally varying sentence series in which the same context contains the same verb in light-verb and full-verb uses. Two probing experiments show that language models differentiate between these uses even in minimal contexts and exhibit separable patterns across object types. We release the dataset, generation code, and materials as a reusable resource. The framework supports extensions to broader contexts, additional verbs, and other languages.

\end{abstract}

\section{Introduction}
\label{sec:intro}

Verbs have different functions in sentences. In a \textit{full-verb construction}, the verb is the main lexical predicate and contributes independent lexical meaning, as in \textit{make a cake}; in a \textit{light-verb construction} (LVC), a semantically weak or ``light'' verb (LV) provides verbal inflection, syntactic realization, and argument-structural support combining with a predicative noun, which contributes the main lexical content as in \textit{make a decision}. 
%
LVCs have long been studied in syntax, lexical semantics, phraseology, typology, and computational linguistics \citep{Jespersen42,GrimshawMester88,Butt95,Butt10,TuRoth11,Nagy-etal20,FleischhauerLatrouite25}, but remain underexploited as diagnostic material for evaluating language models. Existing computational work has largely treated LVCs as an identification or extraction problem: given a candidate verb--noun combination, the task is to determine whether it is an LVC. This has led to supervised and multilingual detection and classification experiments \citep{Vincze-etal13,ChenPalmer15,Vaidya-etal16,CordeiroCandito19}, as well as broader overviews of LVC and verbal-MWE identification \citep{Constant-etal17,Tan-etal21}. Yet identification and controlled probing address different questions: the former asks whether a verb--noun expression is an LVC, whereas the latter asks whether a model distinguishes light and full uses of the same verb under matched syntactic and contextual conditions. This matters because familiar LVCs may be recognized through distributional association, lexical memorization, or noun-category cues, without representation of the lexical-functional contrast itself. Thus, \textit{make a decision} and \textit{make a cake} share the same transitive frame, but only the former realizes an LVC with the predicative nominal base \textit{decision}.

Minimal-pair benchmarks address this gap by isolating linguistic contrasts while holding other sentence properties constant. Targeted syntactic evaluation uses such pairs to probe model representations or test whether models assign higher probability to the expected alternative~\citep{MarvinLinzen18,baroni2026tracing}. BLiMP instantiates this approach for 67 English grammatical phenomena~\citep{Warstadt-etal20};  \citet{jumelet-etal-2026-multiblimp} propose its multilingual extension. SyntaxGym provides a reproducible framework for targeted test-suite experiments based on minimal or near-minimal sentence contrasts~\citep{Gauthier-etal20}, with applications also, e.g., to Chinese~\citep{Xiang-etal21} and Spanish~\citep{TaboasWanner25}. While existing resources mostly contrast acceptable and unacceptable sentences, LVC probing requires contrasting grammatical and natural sentences that differ in the lexical-functional status of the same surface verb.

We present a multilingual dataset for controlled probing of LVCs, currently covering five highly frequent LVs in English, French, and Spanish. It extends the contrastive logic of minimal-pair benchmarks to minimal sentence series: the same sentence context is paired with several objects, yielding LVC readings in half of the cases and full-verb readings in the other half. This design provides more directly comparable items than isolated minimal pairs, improves statistical power, and reduces contextual confounds. Unlike most minimal-pair resources, our dataset contrasts grammatical and natural sentences differing only in the lexical-functional status of the same surface verb. We illustrate its use through contrastive embedding analysis and surprisal-based probing in active and passive configurations, testing whether six models of different families and sizes capture both verb-to-noun and noun-to-verb expectations in LVCs.

\section{Related Work}
\label{sec:related}

NLP has mostly focused on identification and classification of LVCs.
Supervised, syntax-based, lexical-knowledge-based, multilingual, and neural approaches distinguish LVCs from non-LVC verb--noun co-occurrences \citep{TanKanCui06,TuRoth11,Vincze-etal13,ChenPalmer15,Vaidya-etal16,CordeiroCandito19,Nagy-etal20}, often within broader MWE processing and verbal-MWE shared tasks \citep{Constant-etal17,Savary-etal17,Ramisch-etal18}. This work shows that association strength is a central cue for distinguishing conventionalized verb--noun combinations from freer co-occurrences and for classifying them by syntactic and semantic patterns. Corpus-linguistic work on collocation and collostructional analysis provides complementary measures of lexical co-occurrence and word--construction attraction \citep{stefanowitsch2003collostructions,Gries13}. We draw on this tradition to control verb--noun conventionality, while treating LV status as a distinct lexical-functional contrast. Accordingly, lower surprisal for LVCs is interpreted not as frequency-independent categorical recognition, but as graded sensitivity to LVC-related expectancy patterns.

Recent work on MWEs in transformer models further motivates controlled LVC/FVC contrasts: MWE semantics is often captured inconsistently and may rely on surface patterns or memorized associations, while probing work on idiomaticity shows that MWE-related information can appear in contextual representations, but remains entangled with lexical and distributional properties. Surprisal-based studies of idioms and MWEs likewise suggest that conventionalized expressions may be processed more easily because their components are more predictable in context. This motivates our active/passive design: if an LV licenses a conventional nominal base, the noun should receive lower surprisal in active configurations; conversely, if a predicative nominal base licenses a light verb, the verb should receive lower surprisal in passive configurations.

Methodologically, our work builds on targeted minimal-pair evaluation of language models \citep{MarvinLinzen18,Warstadt-etal20,Gauthier-etal20,Xiang-etal21,jumelet-etal-2026-multiblimp,baroni2026tracing}, surprisal-based accounts of incremental processing \citep{Levy08,SmithLevy13,GoodkindBicknell18}, and targeted test-suite approaches using region-level surprisal or probability contrasts \citep{Gauthier-etal20,Hu-etal20,Wilcox-etal21,Beyer-etal21,TaboasWanner25}. Extending this work, our benchmark contrasts matched grammatical sentences, rather than acceptable--unacceptable pairs, to test whether models encode directional expectations between predicative nouns and their LVs.

\section{Dataset Construction}
\label{sec:dataset}

Our seed dataset inventory includes five high-frequency LVs in each of the three languages: English: \textit{make}, \textit{take}, \textit{give}, \textit{have}, and \textit{receive}; French: \textit{faire}, \textit{prendre}, \textit{donner}, \textit{avoir}, and \textit{recevoir}; and Spanish: \textit{hacer}, \textit{tomar}, \textit{dar}, \textit{tener}, and \textit{recibir}. 
We selected the English and French frequently occurring as LVCs in CollFrEn \citep{Fisas-etal20}, while the Spanish verbs were manually selected to preserve the same cross-linguistic verb classes.\footnote{So far, we excluded other frequent English LVC verbs, such as \textit{do} and \textit{get}, and their equivalents in the other languages. These verbs are often included in LVC recognition tasks; cf., e.g., \citep{TuRoth11}. However, many of their transitive uses are generic activity predicates and do not yield clear minimal-pair contrasts between LV and full-verb readings.}


All candidates were validated according to three criteria: the noun had to be a predicative nominal base, the verb had to be semantically light in the construction, and the combination had to instantiate a canonical LV-construction corresponding to either the Oper$_1$ or Oper$_2$ lexical functions in Explanatory Combinatorial Lexicology \citep{Melcuk-etal95}. Borderline cases and idiomatic expressions were excluded.
Based on this LV selection, we constructed a corresponding full-verb--noun set. We selected a set of nouns denoting concrete objects that could occur after each verb.  
 The constructions in the full-verb condition had to be grammatical and natural, instantiate the same broad syntactic frame of the LVC condition, and involve a noun with which the verb expresses an independent lexical meaning. Table~\ref{tab:examples_vn_pairs} exemplifies the resulting contrasts for English; for French and Spanish examples, see Appendix \ref{appendix:subsec:light-full-verbs}.
 
For each verb, a set of natural-language sentence contexts was generated and paired with the verb + object constructions. 
Sentence contexts vary across several grammatical properties, including voice, tense, length, and number of specifications, allowing different degrees of stimulus variation and experimental control; cf. Table \ref{tab:example_sentences} for some examples.\footnote{The dataset, along with the code and materials used for its generation are released with documentation for reuse at \url{https://github.com/XplainLing/LVC_sentences_database}. Details on the dataset creation and composition are provided in Appendix \ref{sec:appendix-data}.}

\begin{table}[t]
\centering
\small
\begin{tabular}{lll}
\hline
Verb & LVC use & Full-verb use \\
\hline
\textit{make} & \textit{make a decision} & \textit{make a cake} \\
\textit{take} & \textit{take a walk} & \textit{take a mug} \\
\textit{have} & \textit{have a celebration} & \textit{have a car} \\
\textit{give} & \textit{give a look} & \textit{give a coin} \\
\textit{receive} & \textit{receive attention} & \textit{receive a package} \\
\hline
\end{tabular}
\caption{Examples of LVC vs. full-verb noun contrasts.}
\label{tab:examples_vn_pairs}
\end{table}


\begin{table}[t]
\centering
\small
\begin{tabular}{ll}
\hline
Condition & Example \\
\hline
LVC-active & \textit{During the day, they made a decision.} \\
Full-active & \textit{During the day, they made a cake.} \\
LVC-passive & \textit{During the day, a decision was made.} \\
Full-passive & \textit{During the day, a cake was made.} \\
\hline
\end{tabular}
\caption{Examples of matched active and passive sentence-level contrasts for English.}
\label{tab:example_sentences}
\end{table}



\section{Experiments and their Results}
\label{sec:exp}
In what follows, we present the results of the evaluation of the proposed dataset, obtained with minimal contexts in experiments carried out with six different models: \texttt{google/gemma-3-270m}  \citep{gemma3technicalreport}, \texttt{google/gemma-2-2b} \citep{gemma2technicalreport}, \texttt{Qwen/Qwen3.5-0.8B} \citep{qwen35}, \texttt{Qwen/Qwen2.5-3B} \citep{qwen25technicalreport}, \texttt{facebook/opt-1.3b} \citep{opttechnicalreport}, and \texttt{EleutherAI/pythia-410m} \citep{pythiasuite}. The models were chosen to span parameter scale, model family, training generation, and language orientation.
 This allows us to ask whether the contrast is reflected robustly in the contextual representations, and whether cross-linguistic separability varies with model type and scale.

We evaluate the proposed dataset along two strands. First, we use it for surprisal-based probing of language models, testing whether models assign different expectations to LVC and full-verb configurations when the surface verb and syntactic frame are controlled. Second, we conduct a contrastive embedding analysis that examines whether object-level contextual representations separate LVC objects from full-verb objects under matched verbal and sentential contexts.
Together, these experiments are intended to assess whether the light/full-verb distinction is reflected in model expectations and model-internal representations.

\subsection{Surprisal-Based Probing}
\label{subsec:surprisal}

For a token $w_t$ in context $w_{<t}$, \textit{surprisal} is defined as the negative log-probability assigned to that token by a language model:
\[
S(w_t) = - \log P(w_t \mid w_{<t}).
\]
We report surprisal in nits. For multi-token critical regions, we use the maximum surprisal score across the tokens.
The surprisal experiment follows the SyntaxGym evaluation format \citep{Gauthier-etal20}: each test item consists of matched conditions and manually defined critical regions. We compute token-level surprisals for the critical regions using autoregressive language models and compare differences across verbs and classes. 
In active voice configurations, the critical region is the noun following the verb. This tests whether the preceding verb and context make an LVC-compatible nominal base more or less expected than a full-verb object. 
In passive voice configurations, the critical region is the verb following the nominal base. This tests whether a predicative nominal base increases the expectation for its conventional LV. For each condition, we compare surprisal values across matched LVC and full-verb sentences.

\begin{figure}[t!]
    \centering
    \includegraphics[width=1\linewidth]{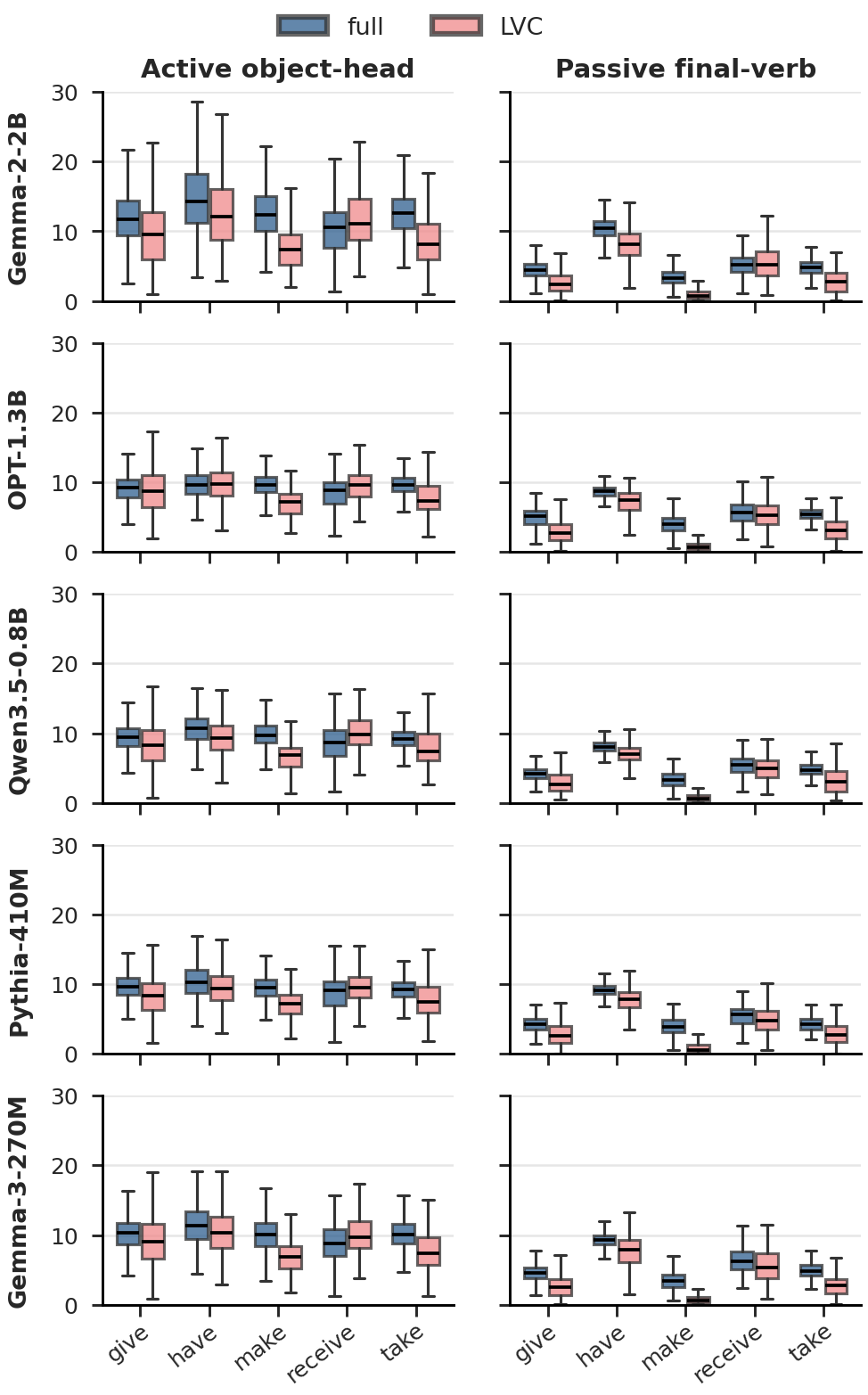}
    \caption{Surprisal (y-axis, in nits) for five language models on 47,600 active and 47,600 passive English sentences. For active sentences, surprisal is measured at the head of the sentence-final object; for passive sentences, it is measured at the sentence-final verb.}
    \label{fig:box_plots_multi_models}
\end{figure}

We compared surprisal between LVCs and full verbs using linear mixed-effects models with random intercepts for sentence context and verb+object. Across languages, models and voice, LVCs showed significantly lower surprisals [Holm-adjusted \textit{p} $\le$ .033]; full results are reported in Appendix~\ref{app:surprisal}. Figure~\ref{fig:box_plots_qwen} shows the contrastive surprisal scores for \texttt{Qwen2.5-3B}, whose patterns are broadly representative across models, in active and passive voice configurations for all three languages; cf. Appendix~\ref{app:surprisal} for the other five models.
In  active voice, the critical regions in the LVC condition receive lower surprisal than their full-verb counterparts, except for \textit{receive} and its French and Spanish equivalents, \textit{recevoir}/\textit{recibir}. This suggests that the model assigns higher contextual probability to nominal bases occurring with conventional LVs such as \textit{give}/\textit{donner}/\textit{dar}, \textit{have}/\textit{avoir}/\textit{tener}, \textit{make}/\textit{faire}/\textit{hacer}, and \textit{take}/\textit{prendre}/\textit{tomar}, while the status of \textit{receive}/\textit{recevoir}/\textit{recibir} is more ambivalent. In  passive voice, where the preceding noun cues prediction of the following verb, lower LVC surprisal suggests partial sensitivity to associations between predicative nominal bases and their conventional LVs. Variation across verbs is greater in passive than active configurations in all three languages: \textit{make} shows the lowest surprisal values, suggesting that it is the most strongly expected light verb in the tested passive configurations, whereas the corresponding pattern is less clear for French and Spanish; conversely, \textit{have} and its translation equivalents consistently show the highest surprisal values. 
Figure~\ref{fig:box_plots_multi_models} compares the behavior of five models for English (cf. Appendix \ref{sec:appendix:cross-model} for French and Spanish). Overall, the models show consistent behavior, with only minor deviations; see, e.g., \texttt{gemma-2-2b} for \textit{have}-constructions.

\begin{figure}[t!]
    \centering
    \includegraphics[width=1\linewidth]{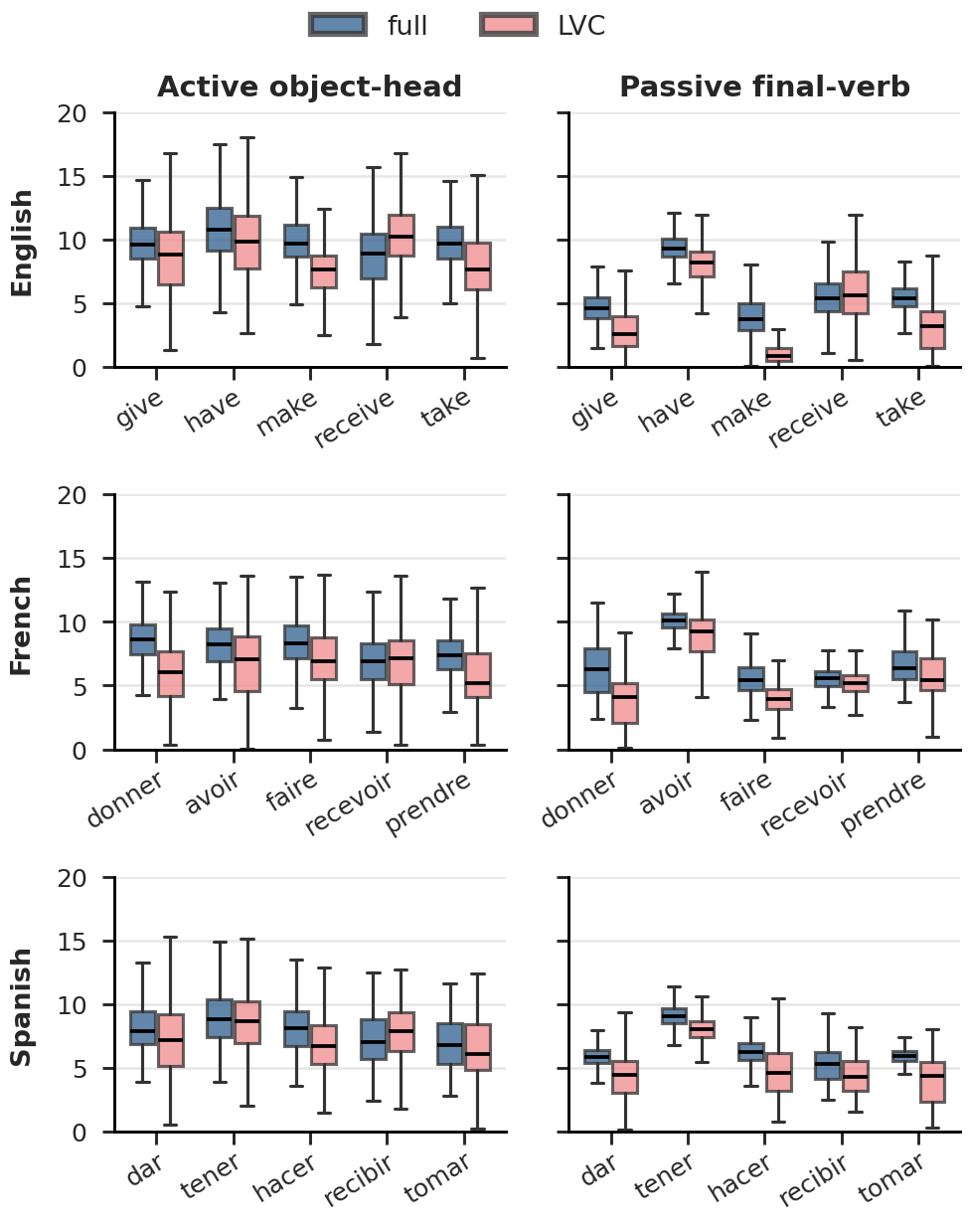}
    \caption{\texttt{Qwen2.5-3B} surprisal (y-axis, in nits) for LVCs and full-verb controls in English, French, and Spanish. The results are computed from 47,600 English, 41,310 French, and 45,080 Spanish sentences per voice. For active sentences, surprisal is measured at the sentence-final object; for passive sentences, at the sentence-final verb.}
    \label{fig:box_plots_qwen}
\end{figure}

\subsection{Contextual Embedding Comparison}
\label{subsec:embeddings}

\begin{figure}[t!]
    \centering
    \includegraphics[width=1\linewidth]{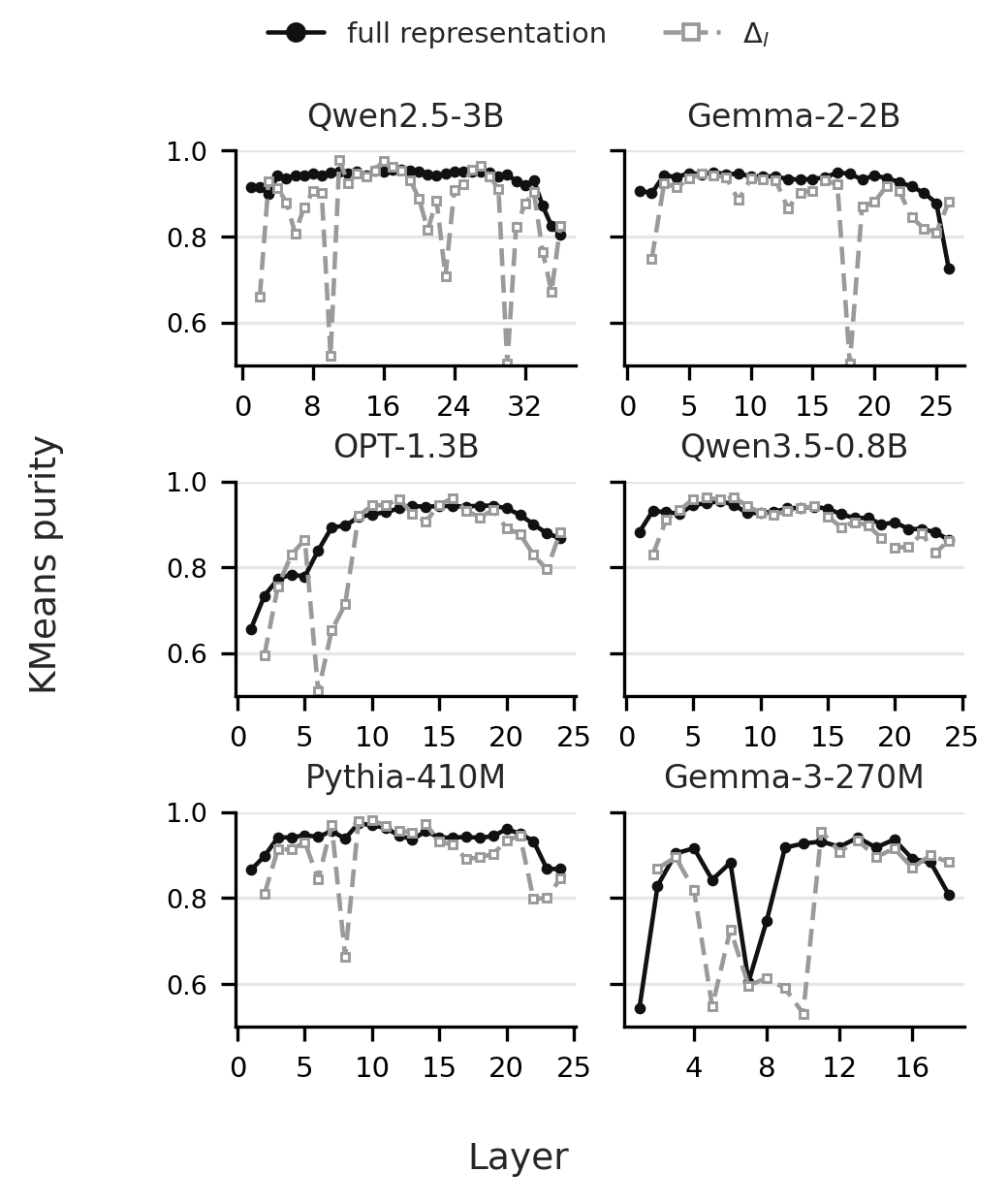}
    \caption{KMeans purity for full object-head hidden states and $\Delta_l$, the change in the object-head representation from layer $l-1$ to layer $l$ for English across all models. Clustering done over 47,600 active sentences.}
    \label{fig:kmeans}
\end{figure}

We compare contextual representations at the object position in LVC and full-verb constructions. For each sentence, we average the hidden states of the object noun phrase, yielding one object-level representation per instance, and test whether these representations separate by construction type. Since the surface verb and sentence template are controlled, such separation indicates that the object representations encode information distinguishing LVC from full-verb contexts, although correlated lexical-semantic properties of the nouns may also contribute.
We assess this by reducing the dimensionality of the object representations and applying clustering analyses, both to the layer-wise representations and to $\Delta_l$, defined as the change from layer $l-1$ to layer $l$. We evaluate whether the induced clusters align with the LVC and full-verb conditions, i.e., whether the surprisal contrast is also reflected in contextual representation space. Figure~\ref{fig:kmeans} illustrates this pattern for English: full object representations show clear light/full-verb separability across most models and layers, though its strength and development vary by model family. Layer-wise changes are less stable, suggesting that the distinction is more consistently reflected in accumulated representations than in individual-layer updates. The multilingual comparison shows further cross-linguistic variation, with some models exhibiting more stable separation across English, French, and Spanish than others; complete results are reported in Appendix~\ref{sec:appendix:cross-model}. Figure~\ref{fig:clusters} provides an example, showing a PCA projection of \texttt{gemma-3-270m} representations in which the two construction types form clearly distinct regions. Together with the surprisal results, these findings show that the light/full-verb contrast is reflected both in model predictions and in contextual representation geometry.

\begin{figure}[h!]
    \centering
    \includegraphics[width=1\linewidth]{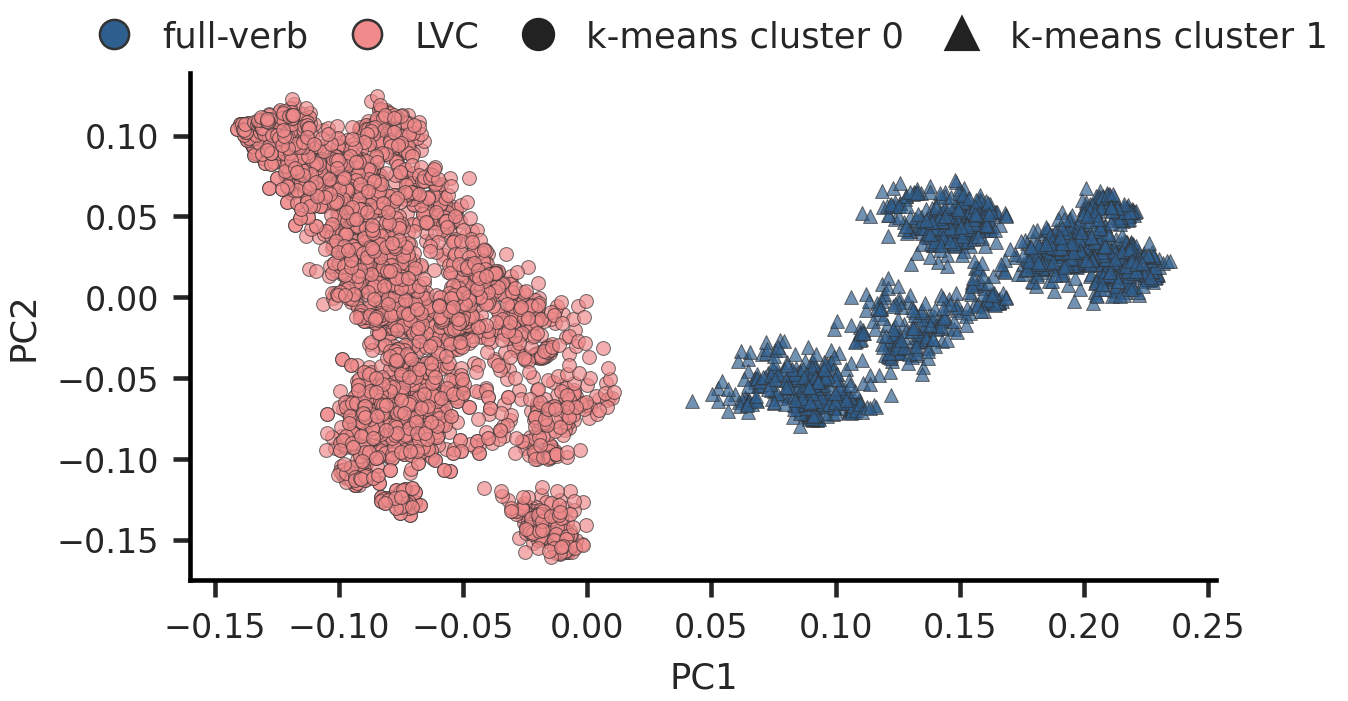}
    \caption{PCA projection of object-level representations from layer 15 of \texttt{gemma-3-270m} for 5,000 active \textit{receive} sentences. Object-token embeddings are averaged per sentence. PC1/PC2 explain 26.6\%/9.0\% of the variance; KMeans uses the first 50 PCs (96.7\% retained variance).
    }
    \label{fig:clusters}
\end{figure}






We further examine whether the observed separation could be explained by lexical properties of the bases themselves, specifically concreteness. 
To this end, we assess its contribution in our English dataset using human-rated concreteness scores on a 1–5 scale, where 5 indicates maximum concreteness~\cite{brysbaert2014concreteness}.
We found that full-verb objects are, on average, more concrete than LVC bases (mean = 4.82, SD = 0.213 vs. mean = 3.00, SD = 0.914). Concreteness thus possibly contributes to the observed separation; however, LVC bases span the full concreteness scale, and embedding-based clustering achieves perfect separation in some layers, beyond what concreteness alone predicts. We therefore interpret the clustering as reflecting semantic differences between the conditions, of which concreteness is an important component, rather than construction type alone. 

\section{Conclusions}
\label{sec:concls}
Our dataset provides a controlled probing instrument for investigating differences between verbs used in LVCs and in full-verb constructions in English, French, and Spanish. It is designed to support further probing experiments, including extensions to additional verbs, broader contextual settings, other languages, and alternative analytical or experimental approaches including psycholinguistic testing. 
The obtained surprisal scores suggest that the six tested models are sensitive to LVCs. LVC-compatible nominal bases are generally less surprising after LVs in active sentences, and predicative nominal bases make their associated LVs more predictable in passive sentences. However, this varies across verbs, which implies the importance of the degree of conventionality of the considered LVCs. The contextual embedding comparison experiment shows that at some point of processing the two object types are clearly separated, suggesting that the models capture at least some of their differential properties. 


\section*{Limitations}
While this database provides controlled minimal pairs, it focuses on a limited set of verbs and constructions, potentially restricting generalization across broader verb classes. Additionally, sentence contexts are semi-controlled, which may reduce ecological validity. 
Finally, our analyses focus on three languages only, leaving a more in-depth cross-linguistic variation in LVCs unexplored. However, we view these choices as enabling precise comparisons, designed this dataset and methods to support future extensions across verbs, contexts, and languages.



\section*{Ethical Statement}
This work introduces a dataset of constructed sentences designed for research on language processing in LLMs. All items have been automatically generated starting from human-designed materials and do not contain offensive or biased content. 

AI support was used for coding and writing assistance, but not for research.


\section*{Acknowledgments}
We are grateful to Beatriz Fisas, Igor Kuzmin and Marco Baroni for their valuable scientific input and insightful discussions.
This work has been funded by the European Research Council (ERC) under the contract No.101201497, granted to the third author. The work reflects the authors’ view only, and the funding agency is not responsible for any use that may be made of the information it contains. 

\bibliography{references}

\appendix




\section{Dataset specifications}
\label{sec:appendix-data}

In this section, we provide the details on the construction of the datasets used in the experiments. In addition to the datasets, we release the full sentence-generation framework and accompanying materials, designed to support large-scale controlled experimentation. The generation pipeline is rule-based and allows users to systematically manipulate a range of grammatical and lexical properties while preserving controlled sentence structure. 

\subsection{Light- and full-verb--object structures}
\label{appendix:subsec:light-full-verbs}

The sentence databases were built around the mentioned five verbs per language, each paired in its light- and full-verb condition with an equal number of nominal objects, with the exception of French \emph{faire}, which contains 35 objects in the full-verb condition. The number of objects in each condition is displayed in Table \ref{tab:number-objects-multiling}. 

\begin{table}[h!]
\centering
\small
\begin{tabular}{lc}
\hline
verb (Engl./Fr./Sp.) & number of objects\\ \hline
\textit{make} / \textit{faire} / \textit{hacer} & 20 / 60 / 50\\
\textit{take} / \textit{prendre} / \textit{tomar} & 20 / 10 / 5\\
\textit{receive} / \textit{recevoir} / \textit{recibir} & 20 / 20 / 5\\
\textit{have} / \textit{avoir} / \textit{tener} & 40 / 24 / 50\\
\textit{give} / \textit{donner} / \textit{dar} & 40 / 20 / 30\\ \hline
\end{tabular}
\caption{\label{tab:number-objects-multiling} The number of objects in each condition across languages.}
\end{table}

Table~\ref{tab:examples_vn_pairs} (Section \ref{sec:dataset})
and Tables~\ref{tab:examples_vn_pairs-fr} and~\ref{tab:examples_vn_pairs-es} below show contrastive examples of the use of these verbs  in LVCs and in full-verb constructions for English, French, and Spanish. 

\begin{table}[h!]
\centering
\small
\begin{tabular}{lll}
\hline
Verb & LVC use & Full-verb use \\
\hline
\textit{faire} & \textit{faire une promesse} & \textit{faire un g\^ateau} \\
\textit{prendre} & \textit{prendre une d\'ecision} & \textit{prendre une tasse} \\
\textit{avoir} & \textit{avoir un accident} & \textit{avoir une voiture} \\
\textit{donner} & \textit{donner un ordre} & \textit{donner un livre} \\
\textit{recevoir} & \textit{recevoir une demande} & \textit{recevoir un fleur} \\
\hline
\end{tabular}
\caption{Examples of contrastive LVC and full-verb -- noun pairs for French.}
\label{tab:examples_vn_pairs-fr}
\end{table}

\begin{table}[h]
\centering
\small
\begin{tabular}{lll}
\hline
Verb & LVC use & Full-verb use \\
\hline
\textit{hacer} & \textit{hacer una prueba} & \textit{hacer una bufanda} \\
\textit{tomar} & \textit{tomar una decisi\'on} & \textit{tomar una botella} \\
\textit{tener} & \textit{tener miedo} & \textit{tener un barco} \\
\textit{dar} & \textit{dar un consejo} & \textit{dar un libro} \\
\textit{recibir} & \textit{recibir una visita} & \textit{recibir un paquete} \\
\hline
\end{tabular}
\caption{Examples of contrastive LVC and full-verb -- noun pairs for Spanish.}
\label{tab:examples_vn_pairs-es}
\end{table}

\subsection{Minimal sentence pairs}
\label{appendix:subsec:sentences}


The light- and full-verb--nominal object structures from above are used to generate minimal sentence pairs. In all generated sentences, the target verb--object construction occurs in sentence-final position, facilitating evaluation with autoregressive language models. For each verb, collocational and full-verb object nouns were paired with temporal sentence contexts and optional adverbial modifiers. Sentences were generated through rule-based templates implementing grammatical constraints on tense, voice, determiner insertion (articles or possessives, where grammatical), adverb placement, and number agreement.

The active sentence database contains sentences with the personal pronoun \textit{they} for English and \textit{ils} for French as subject; for Spanish, the pronouns are dropped to reflect that Spanish is a pro-drop language. Passive forms were generated using manually specified past participles and singular/plural auxiliary agreement. Consider Tables \ref{tab:example_sentences} (Section \ref{sec:dataset}), \ref{tab:example_sentences_fr}, and \ref{tab:example_sentences_es} for illustration of the examples of active and passive sentence structures in the dataset.

\begin{table}[h]
\centering
\small
\begin{tabular}{p{0.04\textwidth}l}
\hline
Cond. & Example \\
\hline
LVC${\rm _a}$ & \textit{Pendant la célébration, ils donneront un concert.} \\
Full${\rm _a}$ & \textit{Pendant la célébration, ils donneront un livre.} \\
LVC${\rm _p}$ & \textit{Pendant la célébration, un concert sera donné.} \\
Full${\rm _p}$ & \textit{Pendant la célébration, un livre sera donné.} \\
\hline
\end{tabular}
\caption{Examples of matched active and passive sentence-level contrasts for French. `Cond.': Condition; `LVC${\rm _a}$': light-verb construction in active; `LVC${\rm _p}$': light-verb construction in passive; `Full${\rm _a}$': full-verb construction in active; `Full${\rm _p}$': full-verb construction in passive.}
\label{tab:example_sentences_fr}
\end{table}

\begin{table}[h]
\centering
\small
\begin{tabular}{ll}
\hline
Cond. & Example \\
\hline
LVC${\rm _a}$ & \textit{Antes de la ceremonia,  darán una respuesta.} \\
Full${\rm _a}$ & \textit{Antes de la ceremonia,  darán una carta.} \\
LVC${\rm _p}$ & \textit{Antes de la ceremonia, una respuesta será dada.} \\
Full${\rm _a}$ & \textit{Antes de la ceremonia, una carta será dada.} \\
\hline
\end{tabular}
\caption{Example of matched active and passive sentence-level contrasts for Spanish. `Cond.': Condition; `LVC${\rm _a}$': light-verb construction in active; `LVC${\rm _p}$': light-verb construction in passive; `Full${\rm _a}$': full-verb construction in active; `Full${\rm _p}$': full-verb construction in passive.}
\label{tab:example_sentences_es}
\end{table}

The dataset includes 280 unique verb+object expressions (244 for French) distributed across collocational and concrete uses, 170 temporal specifications, adverbial modifiers (35 in English, 15 in Spanish and French), 3 tense specifications (English: past simple, future simple, past perfect - Spanish and French: future, present perfect, past perfect) and voice specifications (active, passive). Sentences were automatically inflected for tense and combined with temporal and adverbial contexts. The approximate size of the generated combinations corresponds to over 10,000,000 sentences for English, over 3,000,000 for French and Spanish.




\section{Complementary experimental material}

\subsection{Surprisal analysis}
\label{app:surprisal}
Surprisal differences between LVCs and full-verb controls were assessed separately for each language $\times$ language-model $\times$ voice combination using linear mixed-effects models. Each model included \textit{LVC condition} and \textit{verb} as fixed effects, with crossed random
intercepts for \textit{object} and \textit{sentence context}: $\text{condition} + \text{verb}+ (1 \mid \text{object}) + (1 \mid \text{sentence context})$


Table~\ref{tab:surprisal-condition-effects} summarizes the results. 
Effects are reported as the estimated \textit{LVC $-$ full-verb difference
in surprisal (nits)} with 95\% confidence intervals; negative estimates
indicate lower surprisal for LVCs. $p$-values were Holm-adjusted across
the 36 comparisons.

\begin{table*}[t]
\centering
\small
\begin{tabular}{lllrcc}
\toprule
Language & Model & Voice & $\Delta$ surprisal & 95\% CI & Holm $p$ \\
\midrule

\multicolumn{6}{l}{\textit{English}} \\
& Qwen2.5-3B   & Active  & $-0.974$ & $[-1.475,\,-0.472]$ & .0013 \\
& Qwen2.5-3B   & Passive & $-1.617$ & $[-1.905,\,-1.329]$ & $<.001$ \\
& Gemma-2-2B   & Active  & $-2.293$ & $[-3.174,\,-1.411]$ & $<.001$ \\
& Gemma-2-2B   & Passive & $-1.863$ & $[-2.173,\,-1.552]$ & $<.001$ \\
& OPT-1.3B     & Active  & $-0.634$ & $[-1.104,\,-0.164]$ & .0326 \\
& OPT-1.3B     & Passive & $-1.861$ & $[-2.153,\,-1.569]$ & $<.001$ \\
& Qwen3.5-0.8B & Active  & $-1.040$ & $[-1.567,\,-0.512]$ & .0012 \\
& Qwen3.5-0.8B & Passive & $-1.331$ & $[-1.585,\,-1.078]$ & $<.001$ \\
& Pythia-410M  & Active  & $-1.052$ & $[-1.539,\,-0.565]$ & $<.001$ \\
& Pythia-410M  & Passive & $-1.576$ & $[-1.912,\,-1.240]$ & $<.001$ \\
& Gemma-3-270M & Active  & $-1.179$ & $[-1.789,\,-0.568]$ & .0013 \\
& Gemma-3-270M & Passive & $-1.832$ & $[-2.158,\,-1.505]$ & $<.001$ \\

\midrule
\multicolumn{6}{l}{\textit{French}} \\
& Qwen2.5-3B   & Active  & $-1.337$ & $[-1.887,\,-0.786]$ & $<.001$ \\
& Qwen2.5-3B   & Passive & $-1.401$ & $[-1.765,\,-1.037]$ & $<.001$ \\
& Gemma-2-2B   & Active  & $-2.232$ & $[-2.884,\,-1.581]$ & $<.001$ \\
& Gemma-2-2B   & Passive & $-1.505$ & $[-2.060,\,-0.950]$ & $<.001$ \\
& OPT-1.3B     & Active  & $-0.897$ & $[-1.403,\,-0.392]$ & .0030 \\
& OPT-1.3B     & Passive & $-0.281$ & $[-0.498,\,-0.064]$ & .0326 \\
& Qwen3.5-0.8B & Active  & $-2.050$ & $[-2.534,\,-1.566]$ & $<.001$ \\
& Qwen3.5-0.8B & Passive & $-0.582$ & $[-1.112,\,-0.052]$ & .0326 \\
& Pythia-410M  & Active  & $-0.997$ & $[-1.514,\,-0.480]$ & .0013 \\
& Pythia-410M  & Passive & $-0.728$ & $[-1.017,\,-0.438]$ & $<.001$ \\
& Gemma-3-270M & Active  & $-1.571$ & $[-2.149,\,-0.993]$ & $<.001$ \\
& Gemma-3-270M & Passive & $-0.841$ & $[-1.270,\,-0.413]$ & .0012 \\

\midrule
\multicolumn{6}{l}{\textit{Spanish}} \\
& Qwen2.5-3B   & Active  & $-0.885$ & $[-1.384,\,-0.387]$ & .0030 \\
& Qwen2.5-3B   & Passive & $-1.334$ & $[-1.593,\,-1.076]$ & $<.001$ \\
& Gemma-2-2B   & Active  & $-3.104$ & $[-3.705,\,-2.503]$ & $<.001$ \\
& Gemma-2-2B   & Passive & $-2.062$ & $[-2.403,\,-1.721]$ & $<.001$ \\
& OPT-1.3B     & Active  & $-0.632$ & $[-1.114,\,-0.150]$ & .0326 \\
& OPT-1.3B     & Passive & $-0.800$ & $[-0.998,\,-0.602]$ & $<.001$ \\
& Qwen3.5-0.8B & Active  & $-2.177$ & $[-2.627,\,-1.726]$ & $<.001$ \\
& Qwen3.5-0.8B & Passive & $-1.643$ & $[-1.959,\,-1.327]$ & $<.001$ \\
& Pythia-410M  & Active  & $-0.911$ & $[-1.341,\,-0.480]$ & $<.001$ \\
& Pythia-410M  & Passive & $-0.548$ & $[-0.768,\,-0.329]$ & $<.001$ \\
& Gemma-3-270M & Active  & $-2.671$ & $[-3.220,\,-2.121]$ & $<.001$ \\
& Gemma-3-270M & Passive & $-1.591$ & $[-1.920,\,-1.261]$ & $<.001$ \\

\bottomrule
\end{tabular}

\caption{Mixed-effects model estimates of the surprisal difference
between LVCs and full-verb controls across languages, language models,
and voices. $\Delta$ surprisal denotes the estimated LVC minus
full-control difference in nits; negative values therefore indicate
lower surprisal for LVCs. Confidence intervals are 95\% CIs.
$p$-values are Holm-adjusted across the 36 comparisons.}

\label{tab:surprisal-condition-effects}
\end{table*}

\subsection{Cross-model comparison}
\label{sec:appendix:cross-model}

\begin{figure*}
    \centering
    \includegraphics[width=0.80\linewidth]{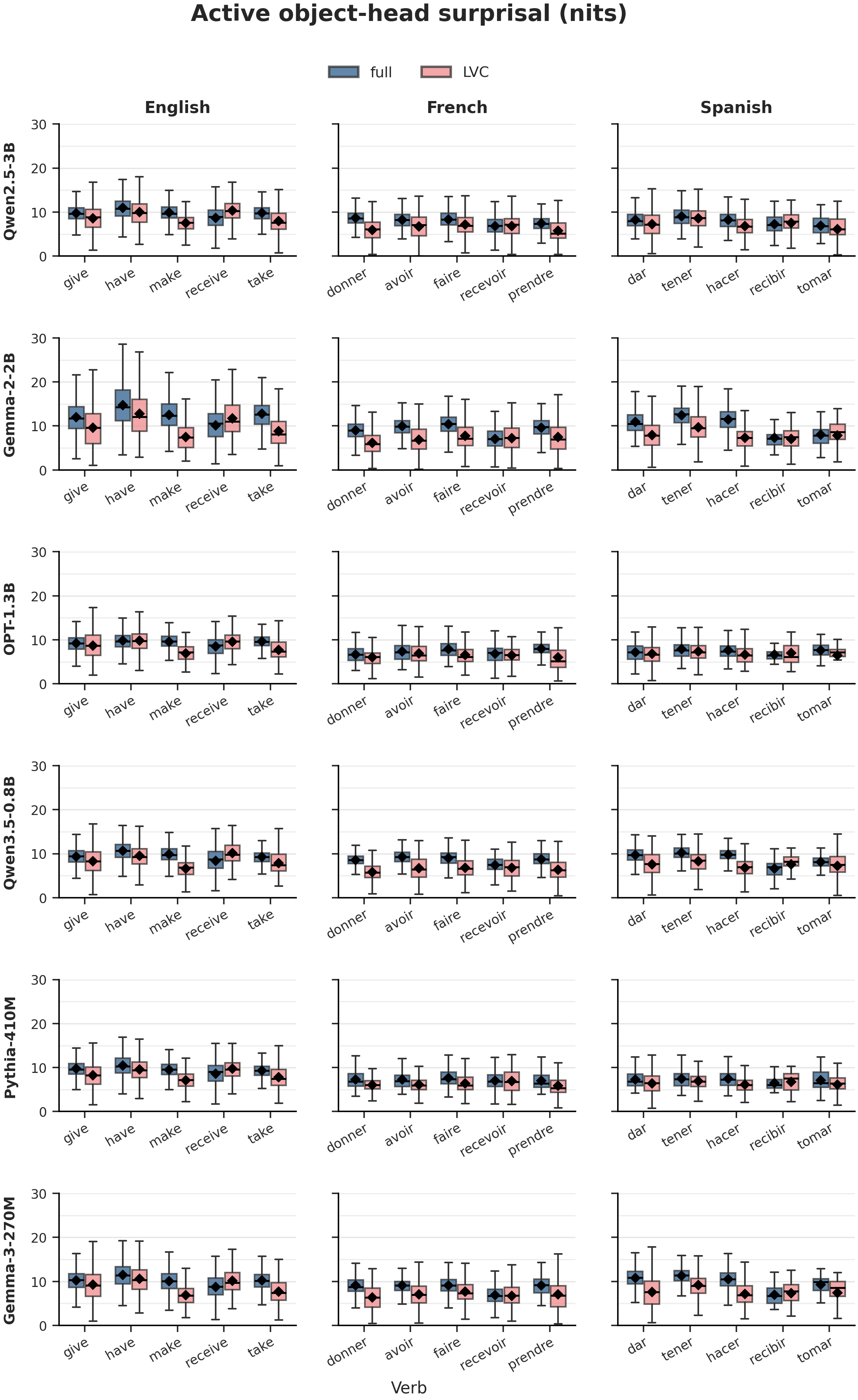}
    \caption{Surprisal (y-axis, in nits) in the six language models for the active voice sentences in all three languages.}
    \label{fig:multilingual1}
\end{figure*}

\begin{figure*}
    \centering
    \includegraphics[width=0.80\linewidth]{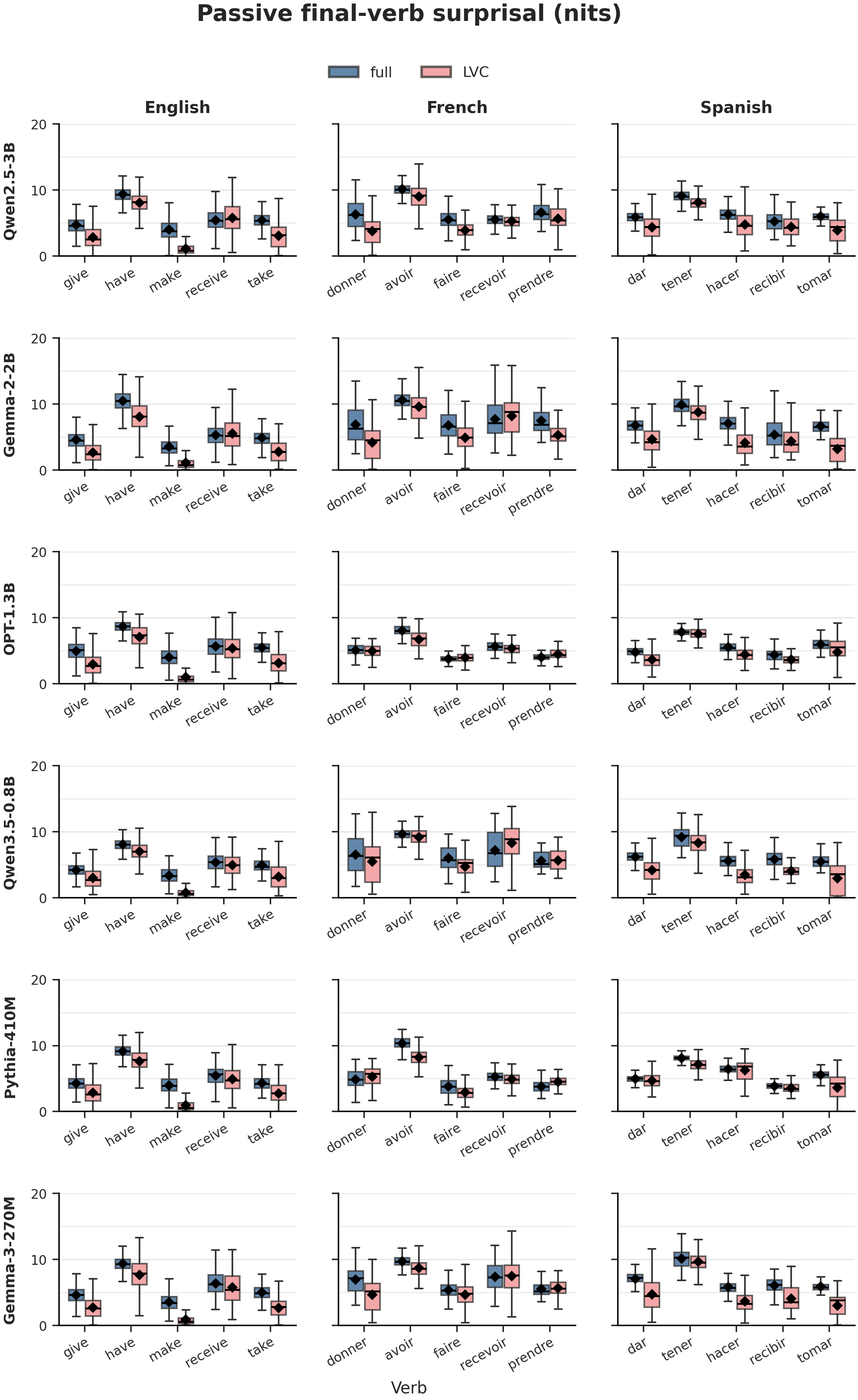}
    \caption{Surprisal (y-axis, in nits) in the six language models for the passive voice sentences in all three languages.}
    \label{fig:multilingual2}
\end{figure*}

Figures \ref{fig:multilingual1} and \ref{fig:multilingual2}
contrast the surprisals of LVC and full-verb constructions in both active and passive voice configurations for English, French, and Spanish as obtained by all six models. In the active voice configuration, the surprisals for English are somewhat higher for all five verbs  across the models, compared to the surprisals for their translation equivalents. This is visible in particular in Gemma2-2B for the verb \textit{have}.

In the passive voice configuration, the surprisal values are more balanced across languages and models. It is interesting to note that despite its high frequency in general language corpora, for \textit{have} and its translation equivalents the values are in this configuration in the average the highest. Further research is needed to determine why this is the case.

\begin{figure*}
    \centering
    \includegraphics[width=0.65\linewidth]{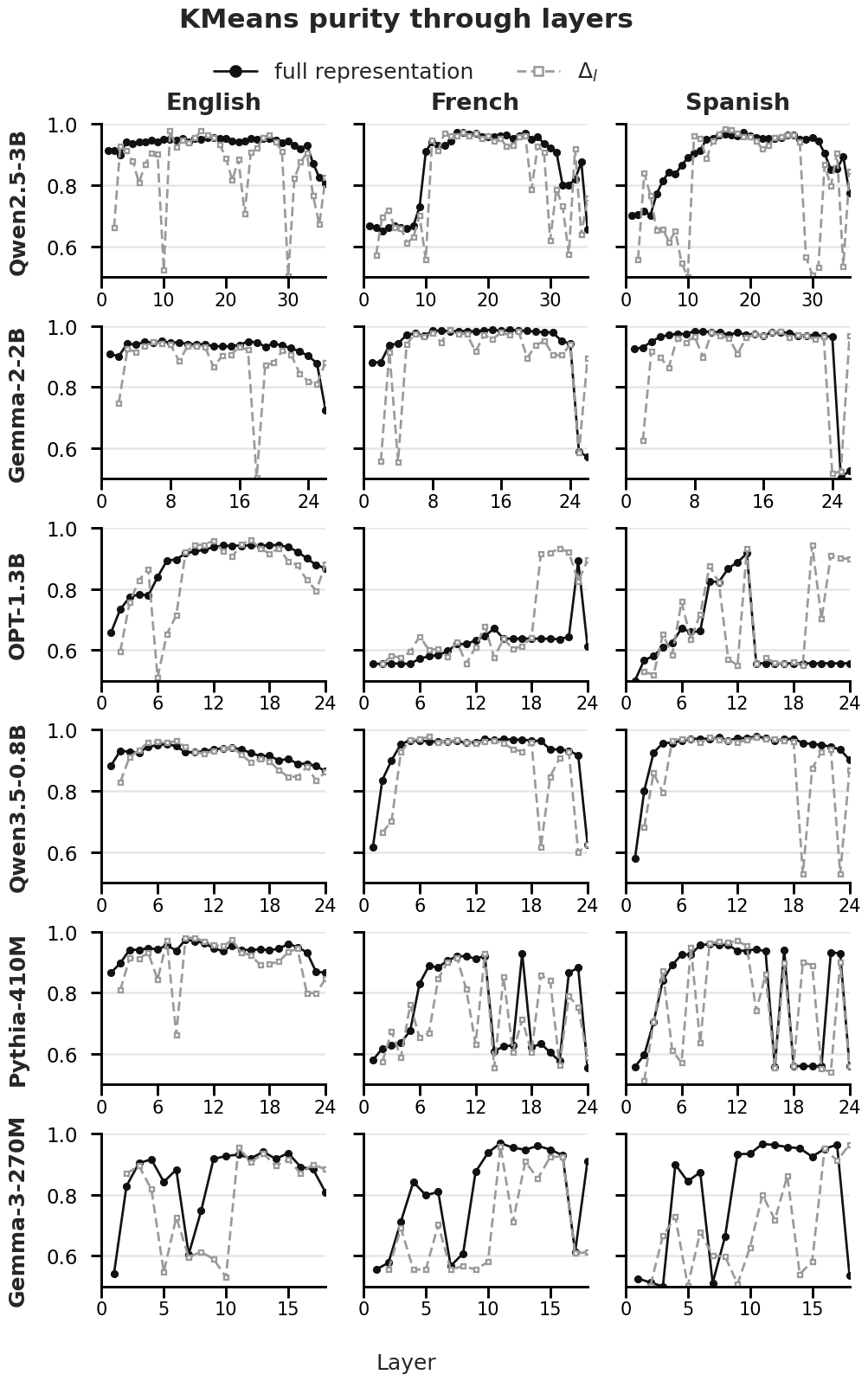}
    \caption{KMeans purity for full object-head hidden states and $\Delta_l$, the change in the object-head representation from layer $l-1$ to layer $l$ for all languages across all models.}
    \label{fig:multilingual3}
\end{figure*}

Figure \ref{fig:multilingual3} contrasts the quality of the separation of the LVCs and full-verb construction hidden states.
\texttt{Qwen2.5-3B}, \texttt{gemma-2-2b}, and \texttt{Qwen3.5-0.8B} show the most stable separation across layers, while the larger \texttt{OPT-1.3b} performs much worse for French and Spanish. At the same time, smaller models such as \texttt{pythia-410m} and \texttt{gemma-3-270m} can still reach very high purity in some layers. Multilingual models, especially the \texttt{Qwen} models, tend to show more consistent results across languages, but multilingual training is not required for this, as the mainly English-trained \texttt{gemma-2-2b} also shows very high separation in French and Spanish. Overall, this suggests that training data and training recipe, including techniques such as knowledge distillation, as well as model generation and architecture, may matter as much as model size or multilingual exposure. The results from \texttt{gemma-2-2b} are specially relevant, since its strong cross-lingual separation may partly reflect its extensive training and distillation from a larger model despite its mainly English-oriented training data. Since these factors vary together across model families, however, we cannot isolate the effect of any one of them from this comparison alone.

\begin{figure*}
    \centering
    \includegraphics[width=\linewidth]{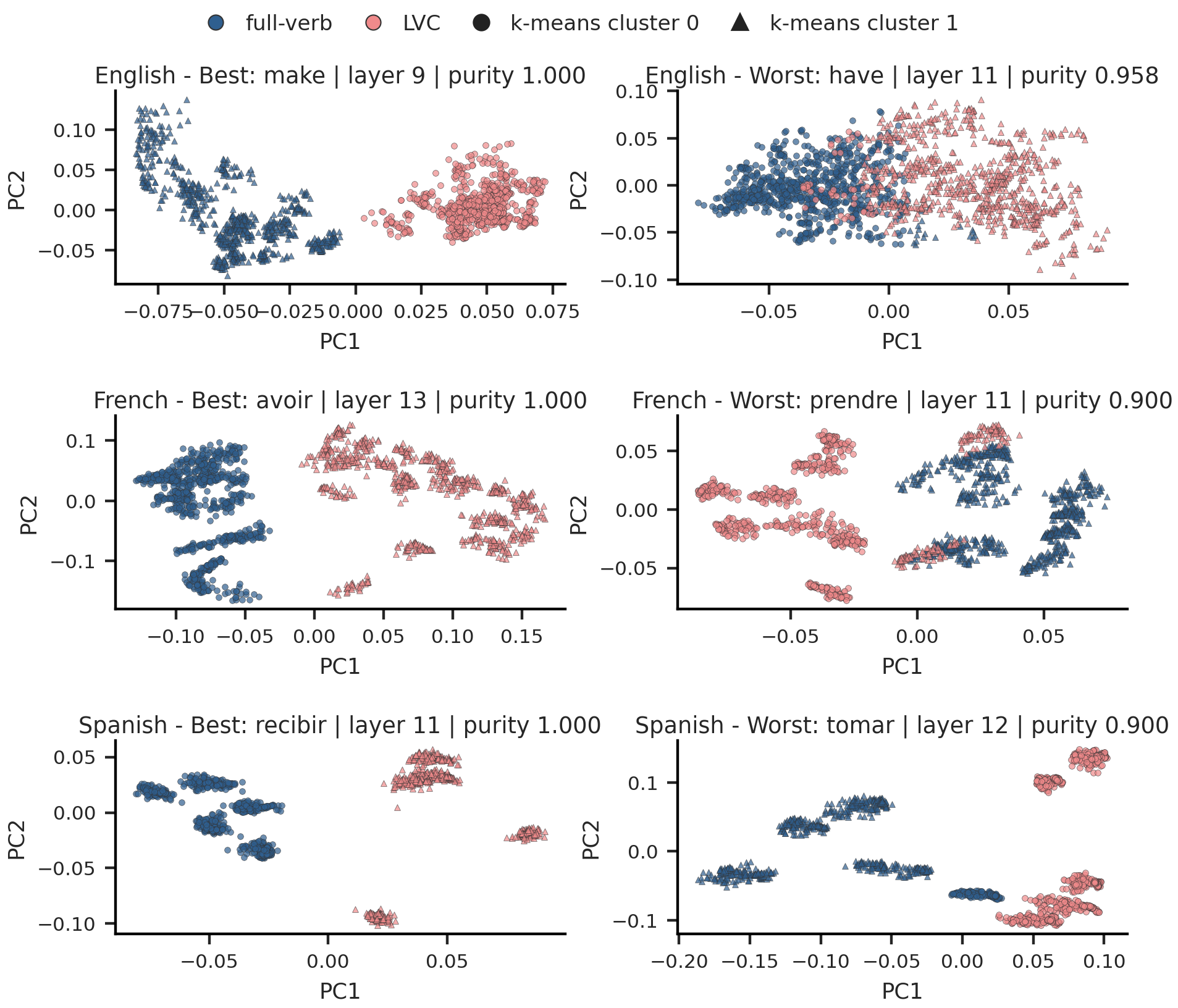}
    \caption{PCA projection of object-level contextual representations from specific layers of \texttt{gemma-3-270m}, computed over 1,610 active sentences. On the representation, each point corresponds to one sentence, with the object representation obtained by averaging the embeddings of all object tokens. KMeans clustering was performed on the first 50 PCA dimensions.}
    \label{fig:separability_layers}
\end{figure*}

Figure~\ref{fig:separability_layers} illustrates PCA projections of object-level contextual representations in \texttt{Gemma-3-270m} for the English, French, and Spanish verbs that yield the lowest and highest cluster purity. Despite the parallels observed across the three languages in terms of surprisal values, the representation-level analysis reveals interesting contrasts. In English, the weakest separation is observed for \textit{have}, whereas its French equivalent, \textit{avoir}, yields the strongest separation. By contrast, French and Spanish coincide in identifying \textit{prendre}/\textit{tomar} as the verbs with the weakest separation.

\end{document}